\documentclass[10pt,twocolumn,letterpaper]{article}

\usepackage{cvpr}
\usepackage{times}
\usepackage{epsfig}
\usepackage{graphicx}
\usepackage{amsmath}
\usepackage{amssymb}
\usepackage[utf8]{inputenc} 
\usepackage[T1]{fontenc}    
\usepackage{url}            
\usepackage{booktabs}       
\usepackage{amsfonts}       
\usepackage{nicefrac}       
\usepackage{microtype}      
\usepackage{xcolor}
\usepackage{wrapfig}	
\usepackage[belowskip=0pt,aboveskip=4pt,font=small]{caption}
\usepackage[belowskip=0pt,aboveskip=0pt,font=small]{subcaption}

\usepackage{enumerate}
\usepackage[olditem,oldenum]{paralist}

\usepackage{alltt}
\usepackage{listings}

\belowdisplayskip \abovedisplayskip

\newlength{\sectionReduceTop}
\newlength{\sectionReduceBot}
\newlength{\subsectionReduceTop}
\newlength{\subsectionReduceBot}
\newlength{\abstractReduceTop}
\newlength{\abstractReduceBot}
\newlength{\captionReduceTop}
\newlength{\captionReduceBot}
\newlength{\subsubsectionReduceTop}
\newlength{\subsubsectionReduceBot}

\newlength{\eqnReduceTop}
\newlength{\eqnReduceBot}

\newlength{\horSkip}
\newlength{\verSkip}

\newlength{\figureHeight}
\usepackage{mysymbols}

\usepackage{cite}

\usepackage[pagebackref=true,breaklinks=true,letterpaper=true,colorlinks,bookmarks=false,citecolor=blue,linkcolor=green]{hyperref}

\newcommand{\model}{\mbox{GVQA}\xspace}

\newcommand{\origDataOld}{VQA v1\xspace}
\newcommand{\cpDataOld}{VQA-CP v1\xspace}

\newcommand{\origDataNew}{VQA v2\xspace}
\newcommand{\cpDataNew}{VQA-CP v2\xspace}

\newcommand{\origData}{VQA \xspace}
\newcommand{\cpData}{VQA-CP \xspace}

\newcommand{\myquote}[1]{\emph{`#1'}}

\cvprfinalcopy 

\def\cvprPaperID{112} 

\ifcvprfinal\fi
\begin{document}

\title{Don't Just Assume; Look and Answer:\\ Overcoming Priors for Visual Question Answering}


\author{Aishwarya Agrawal$^1$\thanks{Work partially done while the first author was interning at Allen Institute for Artificial Intelligence.} , Dhruv Batra$^1$$^,$$^2$, Devi Parikh$^1$$^,$$^2$, Aniruddha Kembhavi$^3$ \\
$^1$Georgia Institute of Technology,
$^2$Facebook AI Research, 
$^3$Allen Institute for Artificial Intelligence\\
\texttt{\{aishwarya, dbatra, parikh\}@gatech.edu, anik@allenai.org}}

\maketitle

\begin{abstract}
\vspace{-10pt}
A number of studies have found that today's Visual Question Answering (VQA) models are heavily driven by superficial correlations 
in the training data and lack sufficient image grounding. 
To encourage development of models geared towards the latter, 
we propose a new setting for VQA where for every question type, train and test sets have different prior distributions of answers.
Specifically, we present new splits of the \origDataOld and \origDataNew datasets, 
which we call Visual Question Answering under Changing Priors (\cpDataOld and \cpDataNew respectively).  
First, we evaluate several existing VQA models under this new setting and show that their performance degrades significantly 
compared to the original VQA setting.
Second, 
we propose a novel Grounded Visual Question Answering model (\model) that contains 
inductive biases and restrictions in the architecture specifically designed to prevent the model from 
`cheating' by primarily relying on priors in the training data. 
Specifically, \model explicitly disentangles the recognition of visual concepts present in the image from the identification of plausible answer space for a given question, enabling the model to more robustly generalize across different distributions of answers.
GVQA is built off an existing VQA model -- Stacked Attention Networks (SAN).
Our experiments demonstrate that GVQA significantly outperforms SAN on both \cpDataOld and \cpDataNew datasets. Interestingly, it also outperforms more powerful VQA models such as Multimodal Compact Bilinear Pooling (MCB) in several cases. GVQA offers strengths complementary to SAN when trained and evaluated on the original \origDataOld and \origDataNew datasets. Finally, GVQA is more transparent and interpretable than existing VQA models.
\end{abstract}

\vspace{-14pt}
\section{Introduction}
\label{sec:intro}

\vspace{-5pt}
Automatically answering questions about visual content is considered to be one of the highest goals of
artificial intelligence. 
Visual Question Answering (VQA) poses a rich set of challenges spanning various domains such as computer vision, 
natural language processing, knowledge representation, and reasoning. 
In the last few years, VQA has received a lot of attention -- 
a number of VQA datasets have been curated \cite{VQA,krishna2016visual,zhu2016visual7w,geman2015visual,malinowski2014multi,gao2015you,ren2015exploring,goyal2016making,YinYang} 
and a variety of deep-learning models have been developed 
\cite{VQA,ChenWCGXN15,DBLP:journals/corr/YangHGDS15,DBLP:journals/corr/XuS15a,DBLP:journals/corr/JiangWPL15,DBLP:journals/corr/AndreasRDK15,DBLP:journals/corr/WangWSHD15,kanan,hieco,DBLP:journals/corr/AndreasRDK16,DBLP:journals/corr/ShihSH15,DBLP:journals/corr/kim15,fukui,han,DBLP:journals/corr/IlievskiYF16,DBLP:journals/corr/WuWSHD15,DBLP:journals/corr/XiongMS16,zhou,saito}. 

\begin{figure}[t]
\centering
\includegraphics[width=\linewidth]{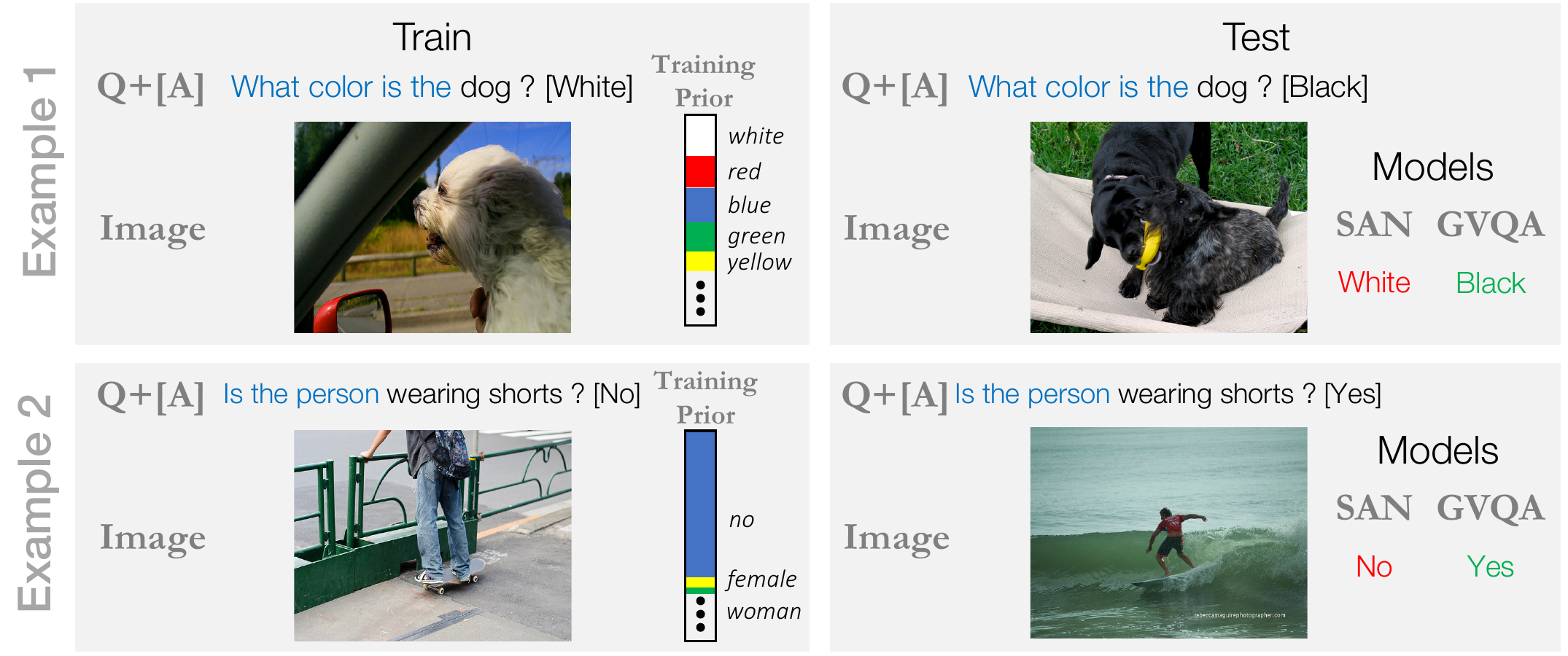}
\caption{ 
Existing VQA models, such as SAN \cite{DBLP:journals/corr/YangHGDS15}, tend to largely rely on strong language priors in train sets, such as, the prior answer (\myquote{white}, \myquote{no}) given the question type (\myquote{what color is the}, \myquote{is the person}). Hence, they suffer significant performance degradation on test image-question pairs whose answers (\myquote{black}, \myquote{yes}) are not amongst the majority answers in train.  
We propose a novel model (\model), built off of SAN that explicitly grounds visual concepts in images, and consequently significantly outperforms SAN in a setting with mismatched priors between train and test.}
\vspace{-4pt}
\label{fig:teaser}
\vspace{-10pt}
\end{figure}

However, a number of studies have found that despite recent progress, 
today's VQA models are heavily driven by superficial correlations  
in the training data and lack sufficient visual grounding \cite{vqa-ba,YinYang,goyal2016making,johnson2016clevr}. It seems that when faced with a difficult learning problem, models typically resort to latching onto the 
language priors in the training data to the point of ignoring the image --  
\eg, overwhelmingly replying to \myquote{how many X?} questions with \myquote{2} (irrespective of $X$), 
\myquote{what color is \ldots?} with \myquote{white}, \myquote{is the \ldots?} with \myquote{yes}. 

One reason for this emergent dissatisfactory behavior is the fundamentally problematic nature of IID train-test splits \emph{in the presence of strong priors}.
As a result, models that intrinsically memorize biases in the training data 
demonstrate acceptable performance on the test set. 
This is problematic for  
benchmarking progress in VQA because it becomes unclear 
what the source of the improvements is -- if models have learned to ground concepts in images or they are driven by memorizing priors in training data.

To help disentangle these factors,  
we present new splits of the \origDataOld \cite{VQA} and \origDataNew \cite{goyal2016making} datasets, called \textbf{Visual Question Answering under Changing Priors} (\textbf{\cpDataOld} and \textbf{\cpDataNew} respectively). 
These new splits are created by re-organizing the train and val splits of the respective VQA datasets in such a way that the distribution of answers 
per question type (\myquote{how many}, \myquote{what color is}, \etc) is by design \emph{different} in the test split compared to the train split (\secref{sec:dataset}). 
One important thing to note: 
we do not change the distribution of the underlying perceptual signals -- the images -- between train and test. 
Generalization across different domains of images (\eg COCO images \vs. web cam images) is an active research area and not the 
focus of this work.  
We change the distribution of \emph{answers for each question type} between train and test. Our hypothesis is that it is reasonable to expect models that are answering questions for the `right reasons' (image grounding) to recognize 
\myquote{black} color at test time even though \myquote{white} is the most popular answer for \myquote{What color is the \ldots?} questions in the train set \figref{fig:teaser}.

To demonstrate the difficulty of our \cpData splits, we report the performance of several existing VQA models \cite{Lu2015, DBLP:journals/corr/AndreasRDK15,DBLP:journals/corr/YangHGDS15,fukui} on these splits. 
Our key finding is that the 
performance of \emph{all tested existing} models drops significantly
when trained and evaluated on the new splits compared to the original splits (\secref{sec:baselines}). This finding provides further confirmation and a novel insight to the growing 
evidence in literature on the behavior of VQA models \cite{vqa-ba,YinYang,goyal2016making,johnson2016clevr}. 

We also propose a novel \textbf{Grounded Visual Question Answering} (\textbf{\model}) model that contains inductive biases and restrictions in the architecture specifically designed to prevent it from `cheating' by primarily relying on priors in the training data (\secref{sec:model}). 
\model is motivated by the intuition that questions in VQA provide two key pieces of information: \\  
(1) What should be recognized? Or what visual concepts in the image need to be reasoned about to answer the question
(\eg, \myquote{What color is the plate?} requires looking at the plate in the image),  \\ 
(2) What should be said? Or what is the space of plausible answers
(\eg, \myquote{What color \ldots?} questions need to be answered with names of colors). 

Our hypothesis is that models that do not explicitly differentiate between these two roles -- 
which is the case for most existing models in literature -- tend to confuse these two signals. They end up learning from question-answer pairs that a 
plausible color of a plate is white, and at test time, rely on this correlation more so than the specific plate in the image the question is about.  
\model explicitly disentangles the visual concept recognition from the answer space prediction. 

GVQA is built off of an existing VQA model -- Stacked Attention Networks (SAN) \cite{DBLP:journals/corr/YangHGDS15}. Our experiments demonstrate that GVQA significantly outperforms SAN on both \cpDataOld and \cpDataNew datasets (\secref{sec:cp_exp}). Interestingly, it also outperforms more powerful VQA models such as Multimodal Compact Bilinear Pooling (MCB) \cite{fukui} in several cases (\secref{sec:cp_exp}). We also show that GVQA offers strengths complementary to SAN when trained and evaluated on the original \origDataOld and \origDataNew datasets (\secref{sec:orig_exp}). Finally, GVQA is more transparent than existing VQA models, in that it produces interpretable intermediate outputs unlike most existing VQA models (\secref{sec:transparency}). 

\vspace{-8pt}
\section{Related Work}
\label{sec:related_work}
\vspace{-7pt}

\textbf{Countering Priors in VQA:} In order to counter the language priors in the \origDataOld dataset, \cite{goyal2016making} balance every question by collecting complementary images for every question. Thus, for every question in the proposed \origDataNew dataset, there are two similar images with different answers to the question. 
By construction, language priors are significantly weaker in the \origDataNew dataset.
However, the train and test distributions are still similar. 
So, leveraging priors from the train set will still benefit the model at test time. \cite{YinYang} balance the yes/no questions on abstract scenes from the \origDataOld dataset in a similar manner. More recently, \cite{kafle_iccv17} propose two new evaluation metrics that compensate for the skewed distribution of question types and for the skewed distribution of 
answers within each question type in the test set. As a remedy for machines using ``shortcuts'' to solve multiple-choice VQA, \cite{DBLP:journals/corr/ChaoHS17} describe several principles for automatic construction of good decoys (the incorrect candidate answers). \cite{wei_lun_cvpr18} study cross-dataset adaptation for VQA. They propose an algorithm for adapting a VQA model trained on one dataset to apply to another dataset with different statistical distribution. All these works indicate that there is an increasing interest in the community to focus on models that are less driven by training priors and are more visually grounded.

\textbf{Compositionality.} Related to the ability to generalize across different answer distributions is the ability to generalize to novel compositions of known concepts learned during training. 
Compositionality has been studied in various forms in the vision community. Zero-shot object recognition using attributes is based on the idea of composing attributes to detect novel object categories \cite{lampert2009learning,jayaraman2014decorrelating}. \cite{atzmon2016learning} have studied compositionality in the domain of image captioning by focusing on structured representations (subject-relation-object triplets). 
\cite{johnson2016clevr} study compositionality in the domain of VQA with synthetic images and questions, with limited vocabulary of objects and attributes.
More recently, \cite{c-vqa} propose a compositional split of the \origDataOld dataset, called C-VQA, that consists of real images and questions (asked by humans) to test the extent to which existing VQA models can answer compositionally novel questions.
However, even in the C-VQA splits, the distribution of answers for each question type does not change much from train to test. Hence, models relying on priors, can still generalize to the test set.

\cite{DBLP:journals/corr/AndreasRDK15,DBLP:journals/corr/AndreasRDK16} have developed Neural Module Networks for VQA that consist of different modules each specialized for a particular task. These modules can be composed together based on the question structure to create a model architecture for the given question. 
We report the performance of this model \cite{DBLP:journals/corr/AndreasRDK15} on our \cpData datasets and find that its performance degrades significantly from the original VQA setting to the proposed CP setting (\secref{sec:baselines}). 

Zero-shot VQA has also been explored in \cite{teney2016zero}. They study a setting for VQA where the test questions (the question string itself or the multiple choices) contain at least one unseen word. \cite{ramakrishnan2017empirical} propose answering questions about unknown objects 
(\eg, \myquote{Is the dog black and white?} where 
\myquote{dog} is never seen in training questions or answers). 
These are orthogonal efforts to our work in that our focus is not in studying if unseen words/concepts can be recognized during testing. We are instead interested in studying the extent to which a model is visually grounded by evaluating its ability to generalize to a different answer distribution for each question type. In our splits, we ensure that concepts seen during test time are present during training to the extent possible.

\vspace{-5pt}
\section{\cpData: Dataset Creation and Analysis}
\label{sec:dataset}
\vspace{-5pt}
The \cpDataOld and \cpDataNew splits are created such that the distribution of answers per question type (\myquote{how many}, \myquote{what color is}, \etc) is different in the test data compared to the training data. These splits are created by re-organizing the training and validation splits of the \origDataOld \cite{VQA} and \origDataNew \cite{goyal2016making} datasets respectively \footnote{We can not use the test splits from \origData datasets because creation of \cpData splits requires access to answer annotations, which are not publicly available on the test sets.}, using the following procedure: 

\textbf{Question Grouping:} Questions having the same question type (first few words of the question -- \myquote{What color is the}, \myquote{What room is}, etc.) and the same ground truth answer are grouped together. For instance, 
\{\myquote{What color is the dog?}, \myquote{white}\} and \{\myquote{What color is the plate?}, \myquote{white}\} are grouped together 
whereas \{\myquote{What color is the dog?}, \myquote{black}\} is put in a different group. 
This grouping is done after merging the QA pairs from the \origData train and val splits. We use the question types provided in the \origData datasets.

\textbf{Greedily Re-splitting:} A greedy approach is used to redistribute data points (image, question, answer) to the \cpData train and test splits so as to maximize the coverage of the \cpData test concepts in the \cpData train split while making sure that questions with the same question type and the same ground truth answer are not repeated between test and train splits. In this procedure, we loop through all the groups created above, and in every iteration, we add the current group to the \cpData test split unless the group has already been assigned to the \cpData train split. We always maintain a set of concepts\footnote{For a given group, concepts are the set of all unique words present in the question type and the ground truth answer belonging to that group.} belonging to the groups in the \cpData test split that have not yet been covered by the groups in the \cpData train split. We then pick the group that covers majority of the concepts in the set, from the groups that have not yet been assigned to either split and add that group to the \cpData train split. We stop when the test split has about 1/3rd the dataset and add the remaining groups (not yet assigned to either split) to the train split.


The above approach results in 98.04\% coverage of test question concepts (set of all unique words in questions after removing stop words -- \myquote{is}, \myquote{are}, \myquote{the}, \etc.) in the train split for \cpDataOld (99.01\% for \cpDataNew), and 95.07\% coverage of test answers by the train split's top 1000 answers for \cpDataOld (95.72\% for \cpDataNew). \cpDataOld\ train consists of $\sim$118K images, $\sim$245K questions and $\sim$2.5M answers ($\sim$121K images, $\sim$438K questions and $\sim$4.4M answers for \cpDataNew train). \cpDataOld\ test consists of $\sim$87K images, $\sim$125K questions and $\sim$1.3M answers ($\sim$98K images, $\sim$220K questions and $\sim$2.2M answers for \cpDataNew test). 



\begin{figure}[t]
\centering
\includegraphics[width=1\linewidth]{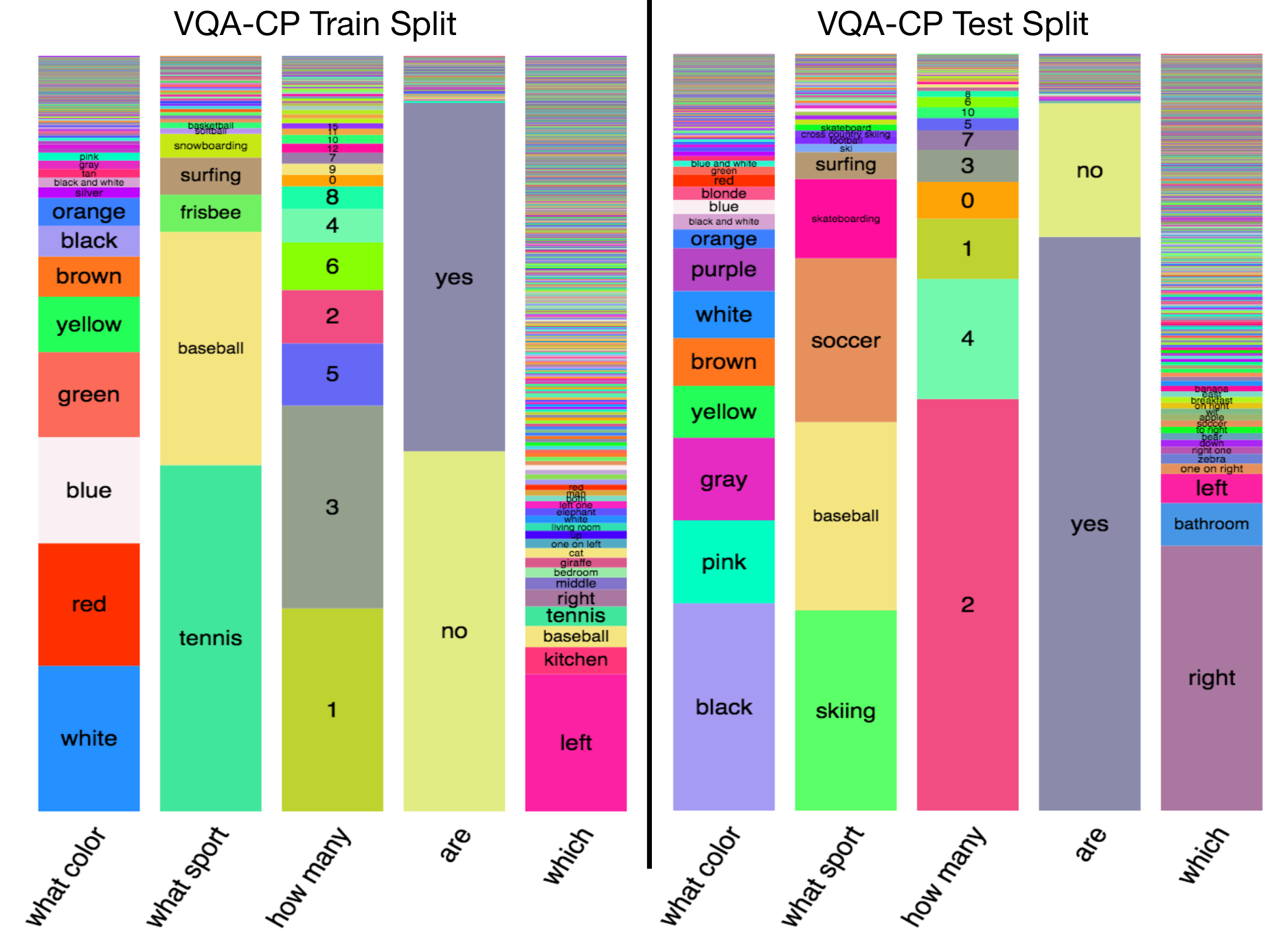}
\caption{Distribution of answers per question type vary significantly between \cpDataOld train (left) and test (right) splits. For instance, \myquote{white} and \myquote{red} are commonly seen answers in train for \myquote{What color}, where as \myquote{black} is the most frequent answer in test. These have been computed for a random sample of 60K questions.}

\vspace{-10pt}
\label{fig:ans_dist}
\setlength{\belowcaptionskip}{-10pt}
\end{figure}

\begin{table*}[h]
\vspace{-2pt}
\setlength{\tabcolsep}{4.5pt}
{\small
\begin{center}
\begin{tabular}{@{}l@{\hspace{4\tabcolsep}}lcccc@{\hspace{4\tabcolsep}}lcccc@{}}
\toprule
Model & Dataset & Overall & Yes/No & Number & Other & Dataset & Overall & Yes/No & Number & Other\\
\midrule
per Q-type prior \cite{VQA} & \origDataOld & 35.13 & 71.31 & 31.93 & 08.86 & \origDataNew & 32.06 & 64.42 & 26.95 & 08.76 \\
    & \cpDataOld & 08.39 & 14.70 & 08.34 & 02.14 & \cpDataNew & 08.76 & 19.36 & 11.70 & 02.39 \\
\midrule
d-LSTM Q \cite{VQA} & \origDataOld & 48.23 & 79.05 & 33.70 & 28.81 & \origDataNew & 43.01 & 67.95 & 30.97 & 27.20 \\
    & \cpDataOld & 20.16 & 35.72 & 11.07 & 08.34 & \cpDataNew & 15.95 & 35.09 & 11.63 & 07.11\\
\midrule
 d-LSTM Q + norm I \cite{Lu2015} & \origDataOld & 54.40 & 79.82 & 33.87 & 40.54 & \origDataNew & 51.61 & 73.06 & 34.41 & 39.85 \\
    & \cpDataOld & 23.51 & 34.53 & 11.40 & 17.42 & \cpDataNew & 19.73 & 34.25 & 11.39 & 14.41 \\
\midrule
NMN \cite{DBLP:journals/corr/AndreasRDK15} & \origDataOld & 54.83 & 80.39 & 33.45 & 41.07 & \origDataNew & 51.62 & 73.38 & 33.23 & 39.93 \\
    & \cpDataOld & 29.64 & 38.85 & 11.23 & 27.88 & \cpDataNew & 27.47 & 38.94 & 11.92 & 25.72 \\
\midrule
SAN \cite{DBLP:journals/corr/YangHGDS15} & \origDataOld & 55.86 & 78.54 & 33.46 & 44.51 & \origDataNew & 52.02 & 68.89 & 34.55 & 43.80 \\
    & \cpDataOld & 26.88 & 35.34 & 11.34 & 24.70 & \cpDataNew & 24.96 & 38.35 & 11.14 & 21.74 \\
\midrule
 MCB \cite{fukui}  & \origDataOld & 60.97 & 81.62 & 34.56 & 52.16 & \origDataNew & 59.71 & 77.91 & 37.47 & 51.76 \\
    & \cpDataOld & 34.39 & 37.96 & 11.80 & 39.90 & \cpDataNew & 36.33 & 41.01 & 11.96 & 40.57 \\
\bottomrule
\end{tabular}
\end{center}
}
\vspace{-10pt}
\caption {We compare the performance of existing VQA models on \cpData test splits (when trained on respective \cpData train splits) to their performance on \origData val splits (when trained on respective \origData train splits). We find that the performance of all tested existing models degrades significantly in the new Changing Priors setting compared to the original VQA setting.}
\vspace{-8pt}
\label{table:baselines}
\end{table*}

\figref{fig:ans_dist} shows the distribution of answers for several question types such as \myquote{what color}, \myquote{what sport}, 
\myquote{how many}, etc. for the train (left) and test (right) splits of the \cpDataOld dataset (see \hyperref[sec:appendix_dataset]{Appendix I} for this analysis of the \cpDataNew dataset). We can see that the distributions of answers for a given question type is significantly different. For instance, \myquote{tennis} is the most frequent answer for the question type \myquote{what sport} in \cpDataOld train split whereas \myquote{skiing} is the most frequent answer for the same question type in \cpDataOld test split. However, for \origDataOld dataset, the distribution for a given question type is similar across train and val splits \cite{VQA} (for instance, \myquote{tennis} is the most frequent answer for both the train and val splits). In the \cpDataOld splits, similar differences can be seen for other question types as well -- \myquote{are}, \myquote{which}.

\vspace{-5pt}
\section{Benchmarking VQA Models on \cpData}
\label{sec:baselines}
\vspace{-7pt}
To demonstrate the difficulty of our \cpData splits, we report the performance of the following baselines and existing VQA models when trained on \cpDataOld and \cpDataNew train splits and evaluated on the corresponding test splits. We compare this with their performance when trained on \origDataOld and \origDataNew train splits and evaluated on the corresponding val splits. Results are presented in \tableref{table:baselines}.

\noindent \textbf{per Q-type prior} \cite{VQA}: Predicting the most popular training answer for the corresponding question type (e.g., \myquote{tennis} for \myquote{What sport \ldots?} questions) \footnote{Note that, ideally the performance of this baseline on \cpData test set should be zero because the answers, given the question type, are different in test and train. But, due to some inter-human disagreement in the datasets, the performance is slightly higher (\tableref{table:baselines}).}. 

\noindent \textbf{Deeper LSTM Question (d-LSTM Q)} \cite{VQA}: Predicting the answer using question alone (``blind'' model). 

\noindent \textbf{Deeper LSTM Question + normalized Image (d-LSTM Q + norm I)} \cite{VQA}: The baseline VQA model.


\noindent \textbf{Neural Module Networks (NMN)} \cite{DBLP:journals/corr/AndreasRDK15}: The model designed to be compositional in nature.


\noindent \textbf{Stacked Attention Networks (SAN)} \cite{DBLP:journals/corr/YangHGDS15}: One of the widely used models for VQA.


\noindent \textbf{Multimodal Compact Bilinear Pooling (MCB)} \cite{fukui}: The winner of the VQA Challenge (on real image) 2016.

\noindent Brief descriptions of all of these models are in \hyperref[sec:appendix_baselines]{Appendix II}.



\begin{figure*}[h]
\centering
\includegraphics[width=.85\linewidth]{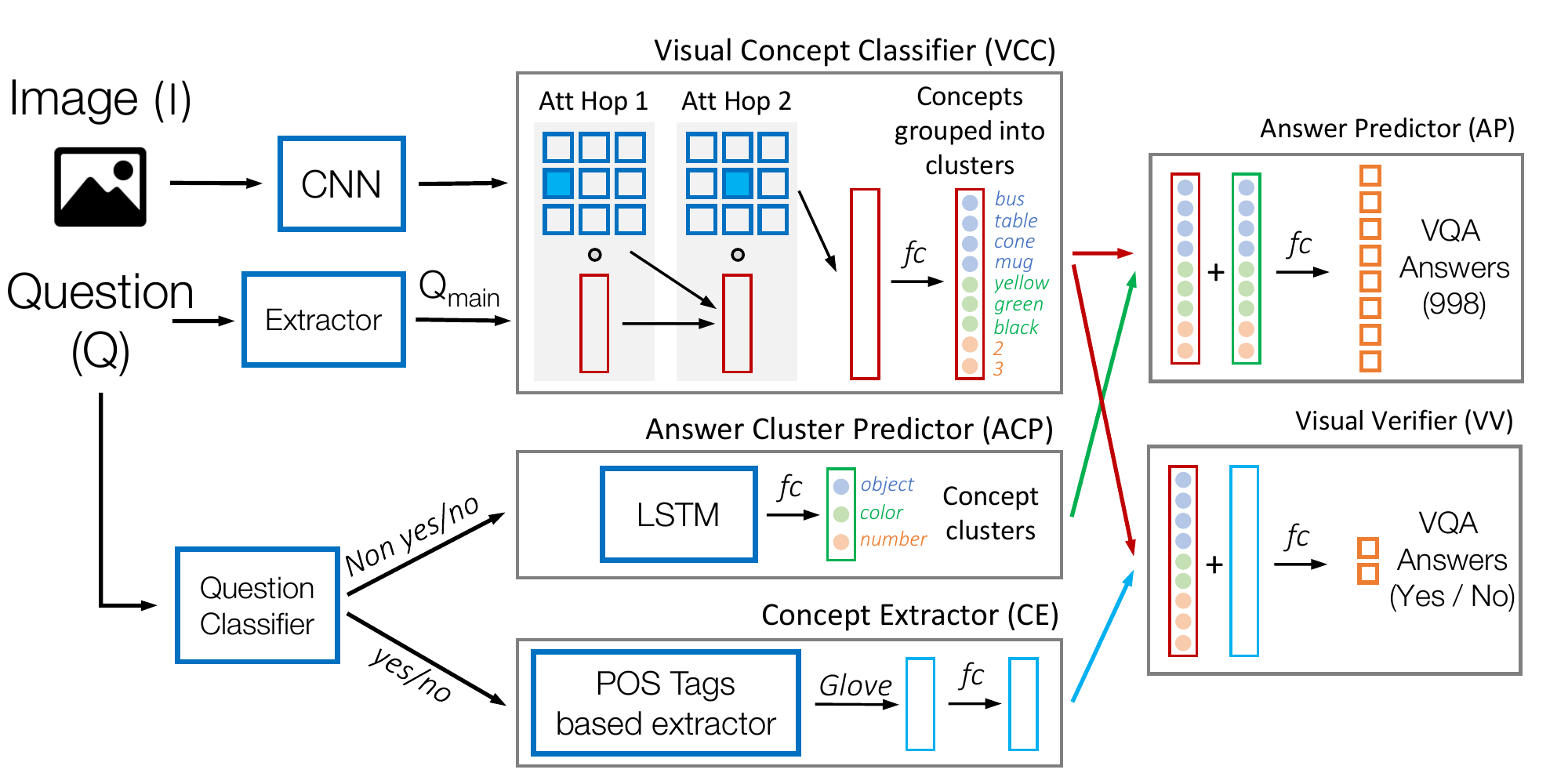}
\caption{The proposed Grounded Visual Question Answering (\model) model. }
\vspace{-5pt}
\label{fig:model}
\setlength{\belowcaptionskip}{-10pt}
\vspace{-5pt}
\end{figure*}

From \tableref{table:baselines}, we can see that the performance of all tested existing VQA models drops significantly in the \cpData setting compared to the original \origData setting. Note that even though the NMN architecture is compositional by design, their performance degrades on the \cpData datasets. We posit this may be because they use an additional LSTM encoding of the question to encode priors in the dataset. 
Also note that the d-LSTM Q + norm I model suffers the largest
drop in overall performance compared to other VQA models, perhaps because other models have more powerful visual processing (for instance, attention on images). Another interesting observation from \tableref{table:baselines} is that the ranking of the models based on overall performance changes from \origData to \cpData. For \origData, SAN outperforms NMN, whereas for \cpData, NMN outperforms SAN. For a brief discussion on trends for different question types, please see \hyperref[sec:appendix_baselines]{Appendix II}.


\vspace{-7pt}
\section{\model model}
\label{sec:model}
\vspace{-7pt}
We now introduce our Grounded Visual Question Answering model (\model). While previous VQA approaches directly map Image-Question tuples $(I,Q)$ to Answers ($A$), \model\ breaks down the task of VQA into two steps: \textbf{Look} - locate the object / image patch needed to answer the question and recognize the visual concepts in the patch, and \textbf{Answer} - identify the space of plausible answers from the question and return the appropriate visual concept from the set of recognized visual concepts by taking into account which concepts are plausible. For instance, when GVQA is asked \myquote{What color is the dog?}, it identifies that the answer should be a color name, locates the patch in the image corresponding to dog, recognizes various visual concepts such as \myquote{black}, \myquote{dog}, \myquote{furry}, and finally outputs the concept \myquote{black} because it is the recognized concept corresponding to color. 
Another novelty in GVQA is that it treats answering yes/no questions as a visual verification task, i.e., it verifies the visual presence/absence of the concept mentioned in the question. For instance, when GVQA is asked \myquote{Is the person wearing shorts?}, it identifies that the concept whose visual presence needs to be verified is \myquote{shorts} 
and answers \myquote{yes} or \myquote{no} depending on whether it recognizes shorts or not in the image (specifically, on the patch corresponding to \myquote{person}).


\model\ is depicted in Figure~\ref{fig:model}. Given a question and an image, the question first goes through the \emph{Question Classifier} and gets classified into yes/no or non yes/no. For non yes/no questions, the GVQA components that get activated are -- 1) \emph{Visual Concept Classifier (VCC)} which takes as input the image features extracted from \emph{CNN} and $Q_{main}$ given by the question \emph{Extractor}, 2) \emph{Answer Cluster Predictor (ACP)} whose input is the entire question. The outputs of \emph{VCC} and \emph{ACP} are fed to the \emph{Answer Predictor (AP)} which produces the answer. For yes/no questions, the GVQA components that get activated are -- 1) \emph{VCC} (similarly to non yes/no), 2) \emph{Concept Extractor (CE)} whose input is the entire question. The outputs of \emph{VCC} and \emph{CE} are fed to the \emph{Visual Verifier (VV)} which predicts \myquote{yes} or \myquote{no}. We present the details of each component below.


\textbf{Visual Concept Classifier (VCC)} is responsible for locating the image patch that is needed to answer the question, as well as producing a set of visual concepts relevant to the located patch. E.g., given \myquote{What is the color of the bus next to the car?}, the VCC is responsible for attending on the bus region and then outputting a set of concepts such as \myquote{bus} and attributes such as its color, count, etc. It consists of a 2-hop attention module based off of Stacked Attention Networks (\cite{DBLP:journals/corr/YangHGDS15}) followed by a stack of binary concept classifiers. The image is fed to the attention module in the form of activations of the last pooling layer of VGG-Net~\cite{VGG}. To prevent the memorization of answer priors per question type, the question is first passed through a language \emph{Extractor}, a simple rule that outputs the string (called $Q_{main}$) after removing the question type substring (eg. \myquote{What kind of}). $Q_{main}$ is embedded using an LSTM and then fed into the attention module. The multi hop attention produces a weighted linear combination of the image region features from VGG-Net, with weights corresponding to the degree of attention for that region. This is followed by a set of fully connected (FC) layers and a stack of $\sim$2000 binary concept classifiers that cover $\sim$95\% of the concepts seen in train. VCC is trained with a binary logistic loss for every concept.
	
The set of VCC concepts is constructed by extracting objects and attributes, pertinent to the answer, from training QA pairs and retaining the most frequent ones. Object concepts are then grouped into a single group where as attribute concepts are clustered into multiple small groups using K-means clustering in Glove embedding space~\cite{Pennington2014GloveGV}, for a total of $C$ clusters.\footnote{We use $C=50$ because it gives better clusters than other values. Also, agglomerative clustering results in similar performance as K-means. More details in \hyperref[sec:appendix_model]{Appendix III}.} 
Concept clustering is required for the purpose of generating negative samples required to train the concept classifiers (for a concept classifier, positive samples are those which contain that concept either in the question or the answer). Since the question does not indicate objects and attributes absent in the image, negative data is generated using the following assumptions: (1) the attended image patch required to answer a question has at most one dominant object in it (2) every object has at most one dominant attribute from each attribute category (e.g., if the color of a bus is red, it can be used as a negative example for all other colors). Given these assumptions, when a concept in a cluster is treated as positive, all other concepts in that cluster are treated as negatives. Note that only a subset of all concept clusters are activated for each question during training, and only these activated clusters contribute to the loss.

\textbf{Question Classifier} classifies the input question $Q$ into 2 categories: Yes-No and non Yes-No using a Glove embedding layer, an LSTM and FC layers. Yes-No questions feed into the CE and the rest feed into the ACP.

\textbf{Answer Cluster Predictor (ACP)} identifies the \textit{type} of the expected answer (\eg object name, color, number, \etc). It is only activated for non yes/no questions. It consists of a Glove embedding layer and an LSTM, followed by FC layers that classify questions into one of the $C$ clusters. The clusters for ACP are created by K-means clustering on (1000) answer classes by embedding each answer in Glove space.\footnote{We first create the clusters for ACP using the answer classes. We then create the clusters for VCC by assigning each VCC concept to one of these ACP clusters using Euclidean distance in Glove embedding space.}

\textbf{Concept Extractor (CE)} extracts question concepts from yes/no questions whose visual presence needs to be verified in the image, using a POS tag based extraction system\footnote{We use NLTK POS tagger. Spacy POS tagger results in similar performance. More details in \hyperref[sec:appendix_model]{Appendix III}.}. E.g., for \myquote{Is the cone green?}, we extract \myquote{green}. The extracted concept is embedded in Glove space followed by FC layers to transform this embedding to the same space as the VCC concepts so that they can be combined by VV. Please see the description of VV below. 

\textbf{Answer Predictor (AP):} Given a set of visual concepts predicted by the VCC, and a concept category predicted by the ACP, the AP's role is to predict the answer. ACP categories correspond to VCC concept clusters (see ACP's and VCC's output classes in \figref{fig:model}. The colors denote the correspondence). Given this alignment, the output of the ACP can be easily mapped into a vector with the same dimensions as the VCC output by simply copying ACP dimensions into positions pertaining to the respective VCC cluster dimensions. The resulting ACP embedding is added element-wise to the VCC embedding followed by FC layers and a softmax activation, yielding a distribution over 998 VQA answer categories (top 1000 training answers minus \myquote{yes} and \myquote{no}). 

\textbf{Visual Verifier (VV):} Given a set of visual concepts predicted by the VCC and the embedding of the concept whose visual presence needs to be verified (given by CE), the VV's role is to verify the presence/absence of the concept in VCC's predictions. Specifically, the CE embedding is added element-wise to the VCC embedding followed by FC layers and a softmax activation, yielding a distribution over two categories -- \myquote{yes} and \myquote{no}.

\textbf{Model Training and Testing:} We first train VCC and ACP on the train split using the cluster labels (for ACP) and visual concept labels (for VCC)\footnote{Note that we do not need additional image labels to train VCC, our labels are extracted automatically from the QA pairs. Same for ACP.}. The inputs to Answer Predictor (and Visual Verifier) are the predictions from VCC and ACP (CE in the case of yes/no questions) on the training data. During training, we use ground truth labels for yes/no and non yes/no questions for the Question Classifier. During testing, we first run the Question Classifier to classify questions into yes/no and non yes/no. And feed the questions into their respective modules to obtain predictions on the test set. Please refer to \hyperref[sec:appendix_model]{Appendix III} for implementation details.

\vspace{-8pt}
\section{Experimental Results}
\label{sec:exp}

\vspace{-8pt}
\subsection{Experiments on \cpDataOld and \cpDataNew}
\label{sec:cp_exp}

\vspace{-5pt}
\begin{table}[t]
\setlength{\tabcolsep}{3pt}
{\small
\begin{center}
\begin{tabular}{@{}llcccc@{}}
\toprule
Dataset & Model & Overall & Yes/No & Number & Other \\
\midrule
\cpDataOld & SAN \cite{DBLP:journals/corr/YangHGDS15} & 26.88 & 35.34 & 11.34 & 24.70 \\
 & \model\ (Ours) & \textbf{39.23} & \textbf{64.72} & \textbf{11.87} & \textbf{24.86} \\
 \midrule
\cpDataNew & SAN \cite{DBLP:journals/corr/YangHGDS15} & 24.96 & 38.35 & 11.14 & 21.74 \\
 & \model\ (Ours) & \textbf{31.30} & \textbf{57.99} & \textbf{13.68} & \textbf{22.14} \\
\bottomrule
\end{tabular}
\end{center}
}
\vspace{-10pt}
\caption {Performance of GVQA (our model) compared to SAN on \cpData datasets. GVQA consistently outperforms SAN.}
\vspace{-15pt}
\label{table:model_acc}
\end{table}
\vspace{-3pt}

\noindent \textbf{Model accuracies:} \tableref{table:model_acc} shows the performance of our \model\ model in comparison to SAN (the model which GVQA is built off of) on \cpDataOld and \cpDataNew datasets using the VQA evaluation metric \cite{VQA}. Accuracies are presented broken down into Yes/No, Number and Other categories.
As it can be seen from \tableref{table:model_acc}, the proposed architectural improvements (in \model) over SAN show a significant boost in the overall performance for both the \cpDataOld (12.35\%) and \cpDataNew (6.34\%) datasets. It is worth noting that owing to the modular nature of the \model\ architecture, one may easily swap in other attention modules into the VCC. 
Interestingly, on the \cpDataOld dataset, \model also outperforms MCB \cite{fukui} and NMN \cite{DBLP:journals/corr/AndreasRDK15} (\tableref{table:baselines}) on the overall metric (mainly for yes/no questions), in spite of being built off of a relatively simpler attention module from SAN, and using relatively less powerful image features (VGG-16) as compared to ResNet-152 being used in MCB. On the \cpDataNew dataset, \model outperforms NMN in overall metric (as well as for number questions) and MCB for yes/no and number questions.

To check if our particular VQA-CP split was causing some irregularities in performance, we created four sets of VQA-CP v2 splits with different random seeds. This also led to a large portion of the dataset (84\%) being covered across the test splits. The results show that GVQA consistently outperforms SAN across all four splits with average improvement being 7.14\% (standard error: 1.36). Please see \hyperref[sec:appendix_additional_splits]{Appendix IV} for performance on each split.

\noindent \textbf{Performance of Model Components}
\textit{Question Classifier}: On the \cpDataOld test set, the LSTM based question classifier obtains 99.84\% accuracy. 
\textit{ACP}: The Top-1 test accuracy is 54.06\%, with 84.25\% for questions whose answers are in attribute clusters and 43.17\% for questions whose answers are in object clusters. The Top-3 accuracy rises to 65.33\%. Note that these accuracies are computed using the automatically created clusters. 
\textit{VCC}: The weighted mean test F1 score across all classifiers is 0.53. The individual concepts are weighted as per the number of positive samples, reflecting the coverage of that concept in the test set. Please refer to \hyperref[sec:appendix_model_components]{Appendix V} for accuracies on the \cpDataNew dataset.

\vspace{-5pt}
\subsection{Role of GVQA Components}
\label{sec:gvqa_components}
\vspace{-5pt}

In order to evaluate the role of various GVQA components, we report the experimental results (on \cpDataOld) by replacing each component in GVQA (denoted by ``- <component>'') with its traditional counterpart, i.e., modules used in traditional VQA models (denoted by `` + <traditional counterpart>''). For instance, GVQA - CE + LSTM represents a model where CE in GVQA has been replaced with an LSTM. The results are presented in \tableref{table:ablation} along with the result of the full GVQA model for reference.

\textbf{GVQA - $Q_{main}$ + $Q_{full}$:} \model's performance when the entire question ($Q_{full}$) is fed into VCC (as opposed to after removing the question type ($Q_{main}$)) is 33.55\% (overall), which is 5.68\% (absolute) less than that with $Q_{main}$. Note that even with feeding the entire question, GVQA outperforms SAN, 
thus demonstrating that removing question type information helps but isn’t the main factor behind the better performance of GVQA. As an additional check, we trained a version of SAN where the input is $Q_{main}$ instead of $Q_{full}$. Results on VQA-CP v2 show that this version of SAN performs 1.36\% better than the original SAN, however still 4.98\% worse than GVQA (with $Q_{main}$). Please see \hyperref[sec:appendix_san_with_qmain]{Appendix VI} for detailed performance of this version of SAN.

\textbf{GVQA - CE + LSTM:} We replace CE with an LSTM (which is trained end-to-end with the Visual Verifier (VV) using VQA loss). The overall performance drops by 11.95\%, with a drop of 28.76\% for yes/no questions. This is an expected result, given that \tableref{table:model_acc} shows that GVQA significantly outperforms SAN on yes/no questions and the CE is a crucial component of the yes/no pipeline. 

\textbf{GVQA - ACP + LSTM:} We replace ACP with an LSTM (which is trained end-to-end with the Answer Predictor (AP) using VQA loss). The overall performance is similar to GVQA. 
But, the presence of ACP makes GVQA transparent and interpretable (see \secref{sec:transparency}).

\textbf{GVQA - VCC$_{loss}$:} We remove the VCC loss and treat the output layer of VCC as an intermediate layer whose activations are passed to the Answer Predictor (AP) and trained end-to-end with AP using VQA loss. The overall performance improves by 1.72\% with biggest improvement in the performance on other questions (3.19\%). This suggests that introducing the visual concept (semantic) loss in between the model pipeline hurts. Although removing VCC loss and training end-to-end with VQA loss achieves better performance, the model is no longer transparent (see \secref{sec:transparency}). Using VCC loss or not is a design choice one would make based on the desired accuracy vs. interpretability trade off.

\textbf{GVQA - VCC$_{loss}$ - ACP + LSTM:} Replacing ACP with an LSTM on top of \textbf{GVQA - VCC$_{loss}$} hurts the overall performance by 2.09\% with biggest drop (4.94\%) for ``other'' questions (see \textbf{GVQA - VCC$_{loss}$} and \textbf{GVQA - VCC$_{loss}$ - ACP + LSTM} rows in \tableref{table:ablation}). This suggests that ACP helps significantly (as compared to an LSTM) in the absence of VCC loss (and it performs similar to an LSTM in the presence of VCC loss, as seen above). In addition, ACP adds interpretability to GVQA. 

\begin{table}[t]
\setlength{\tabcolsep}{2pt}
{\small
\begin{center}
\begin{tabular}{@{}llccc@{}}
\toprule
Model & Overall & Yes/No & Number & Other \\
\midrule
GVQA - $Q_{main}$ + $Q_{full}$ & 33.55 & 51.64 & 11.51 & 24.43 \\
GVQA - CE + LSTM & 27.28 & 35.96 & 11.88 & 24.85 \\
GVQA - ACP + LSTM & 39.40 & 64.72 & 11.73 & 25.33 \\
GVQA - VCC$_{loss}$ & 40.95 & 65.50 & 12.32 & 28.05 \\
GVQA - VCC$_{loss}$ - ACP + LSTM & 38.86 & 65.73 & 11.58 & 23.11 \\
GVQA  & 39.23 & 64.72 & 11.87 & 24.86 \\
\bottomrule
\end{tabular}
\end{center}
}
\vspace{-10pt}
\caption {Experimental results when each component in GVQA (denoted by ``- <component>'') is replaced with its corresponding traditional counterpart (denoted by `` + <traditional counterpart>'').}
\label{table:ablation}
\end{table}

\vspace{-5pt}
\subsection{Experiments on \origDataOld and \origDataNew}
\label{sec:orig_exp}

\begin{table}[t]
\setlength{\tabcolsep}{4.5pt}
{\small
\begin{center}
\begin{tabular}{@{}lcc@{}}
\toprule
Model & \origDataOld & \origDataNew  \\
\midrule
SAN & 55.86 & 52.02 \\
GVQA & 51.12 & 48.24  \\
\midrule
Ensemble (SAN, SAN) & 56.56 & 52.45 \\
Ensemble (GVQA, SAN) & 56.91 & 52.96 \\
\midrule
Oracle (SAN, SAN) & 60.85 & 56.68 \\
Oracle (GVQA, SAN) & 63.77 & 61.96 \\
\bottomrule
\end{tabular}
\end{center}
}
\vspace{-10pt}
\caption {Results of GVQA and SAN on \origDataOld and \origDataNew when trained on the corresponding train splits.}
\vspace{-10pt}
\label{table:vqa_orig}
\end{table}

\vspace{-5pt}
We also trained and evaluated \model on train and val splits of the \origDataOld \cite{VQA} and \origDataNew \cite{goyal2016making} datasets (results in \tableref{table:vqa_orig}\footnote{We present overall and yes/no accuracies only. Please refer to \tableref{table:appendix_vqa_orig} for performance on number and other categories.}). On \origDataOld, \model achieves 51.12\% overall accuracy, which is 4.74\% (absolute) less than SAN. This gap is not surprising because \origDataOld has well-established heavy language priors that existing models (including SAN) can ``memorize'' from train set and exploit on the test set (since test set contains same priors as train set), whereas \model is designed not to. As vision improves, grounded models like GVQA may show improved performance over models that leverage priors from training data. Moreover, it is important to note that the gain (\model acc - SAN acc) on \cpDataOld (12.35\% absolute) is much higher than the loss (SAN acc - \model acc) on \origDataOld (4.74\% absolute).

On \origDataNew, \model under performs SAN by 3.78\% overall, which is less than SAN acc - \model acc on \origDataOld. And it outperforms SAN by 3.14\% for yes/no questions. This shows that when the priors are weaker (in \origDataNew compared to those in \origDataOld), the gap between GVQA and SAN's performance decreases. We also trained and evaluated \model - VCC$_{loss}$ on both the \origDataOld and \origDataNew datasets and found that it performs worse than \model on \origDataOld and similar to GVQA on \origDataNew. So in addition to interpretability, \model is overall better than \model - VCC$_{loss}$ on these original VQA splits.

In order to check whether GVQA has strengths complementary to SAN, we computed the oracle of SAN's and GVQA's performance -- \textbf{Oracle (GVQA, SAN)}, i.e., we pick the predictions of the model with higher accuracy for each test instance. As it can be seen from \tableref{table:vqa_orig}, the Oracle (GVQA, SAN)'s overall performance is 7.91\% higher than that of SAN for \origDataOld (9.94\% for \origDataNew) suggesting that GVQA and SAN have complementary strengths. Also, note that Oracle (\model, SAN) is higher than Oracle (SAN, SAN) for both \origDataOld and \origDataNew, suggesting that GVQA's complementary strengths are more than that of another SAN model (with a different random initialization).

Inspired by this, we report the performance of the ensemble of GVQA and SAN \textbf{Ensemble (GVQA, SAN)} in \tableref{table:vqa_orig}, where the ensemble combines the outputs from the two models using product of confidences of each model. We can see that Ensemble (GVQA, SAN) outperforms Ensemble (SAN, SAN) by 0.35\% overall for \origDataOld (and by 0.51\% for \origDataNew). It is especially better for yes/no questions. We also found that the ensemble of \model - VCC$_{loss}$ with SAN performs worse than Ensemble (SAN, SAN) for both the \origData datasets (refer to \hyperref[sec:appendix_gvqa_without_vcc_results]{Appendix VII} for accuracies). Hence, \model is a better complement of SAN than \model - VCC$_{loss}$, in addition to being more transparent.

\vspace{-5pt}
\subsection{Transparency}
\label{sec:transparency}
\vspace{-5pt}

\begin{figure}[t]
\centering
\includegraphics[width=0.85\linewidth]{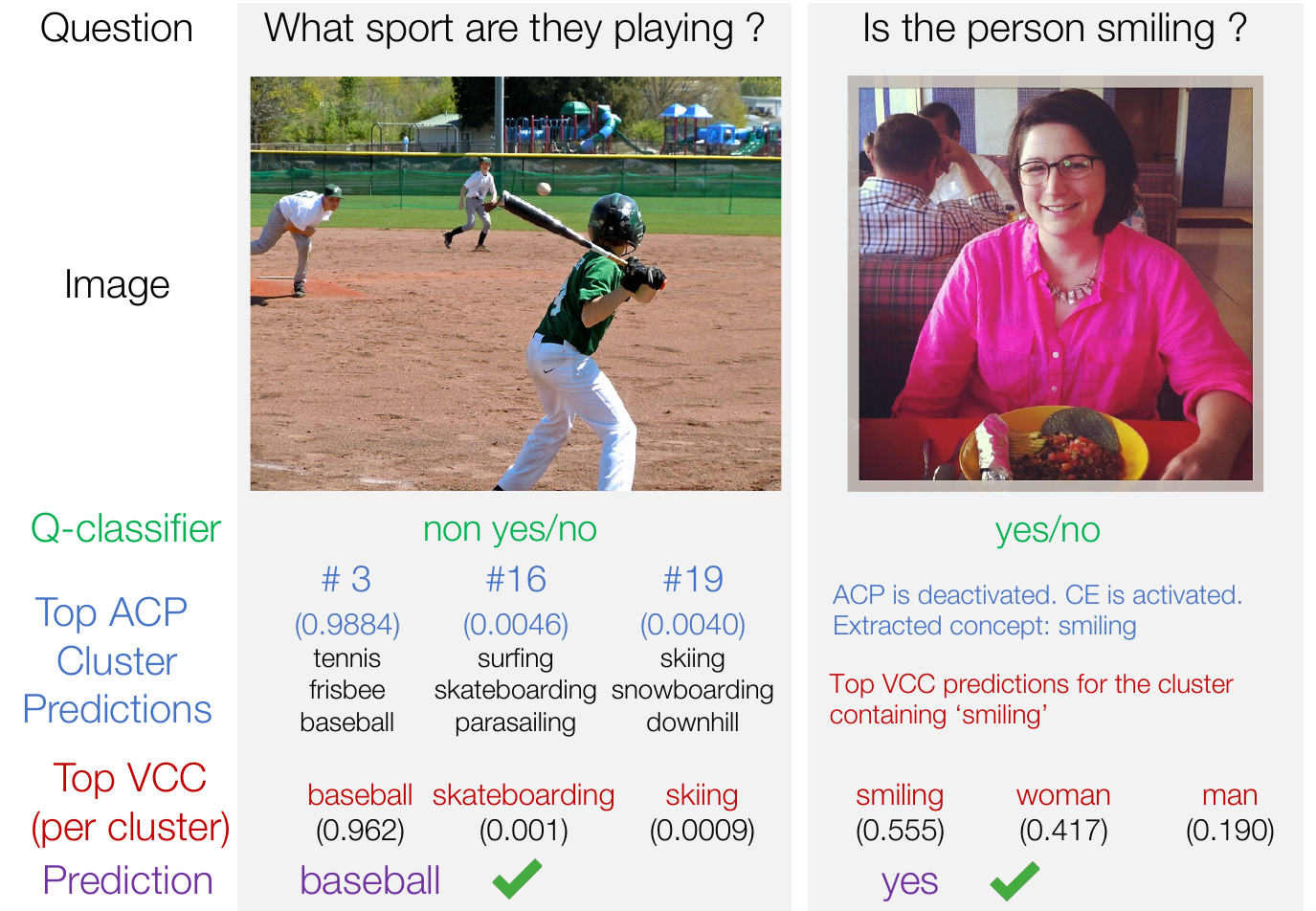}
\caption{Qualitative examples from \model. \textbf{Left:} We show top three answer cluster predictions (along with random concepts from each cluster) by ACP. Corresponding to each cluster predicted by ACP, we show the top visual concept predicted by VCC.
Given these ACP and VCC predictions, the Answer Predictor (AP) predicts the correct answer \myquote{baseball}. \textbf{Right:} Smiling is the concept extracted by the CE whose visual presence in VCC's predictions is verified by the Visual Verifier, resulting in \myquote{yes} as the final answer.}
\label{fig:qual1}
\end{figure}

\begin{figure}[t]
\vspace{-5pt}
\centering
\includegraphics[width=0.8\linewidth]{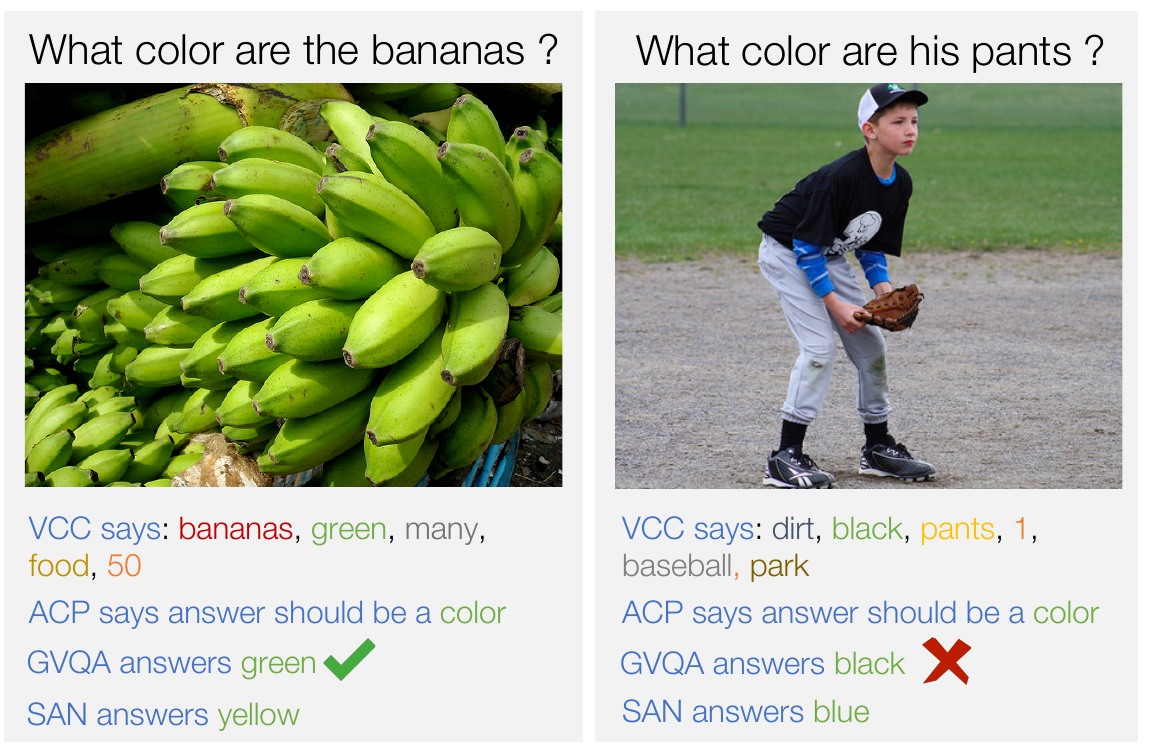}
\caption{\textbf{Left}: \model's prediction (\myquote{green}) can be explained as follows -- ACP predicts that the answer should be a \emph{color}. Of the various visual concepts predicted by VCC, the only concept that is about color is \emph{green}. Hence, \model's output is \myquote{green}.
SAN incorrectly predicts \myquote{yellow}. SAN's architecture doesn't facilitate producing an explanation of why it predicted what it predicted, unlike \model. \textbf{Right}: Both GVQA and SAN incorrectly answer the question. GVQA is incorrect perhaps because VCC predicts \myquote{black}, instead of \myquote{gray}. In order to dig further into why VCC's prediction is incorrect, we can look at the attention map (in \figref{fig:att_map}), which shows that the attention is on the pants for the person's left leg, but on the socks (black in color) for the person's right leg. So, perhaps, VCC's ``black'' prediction is based on the attention on the person's right leg.}
\label{fig:qual2}
\vspace{-10pt}
\end{figure}

The architecture design of \model makes it more transparent than existing VQA models because it produces interpretable intermediate outputs (the outputs of VCC, ACP and the concept string extracted by the CE) unlike most existing VQA models. We show some example predictions from \model in \figref{fig:qual1}. We can see that the intermediate outputs provide insights into why GVQA is predicting what it is predicting and hence enable a system designer to identify the causes of error. This is not easy to do in existing VQA models. \figref{fig:qual2} shows two other examples (one success and one failure) comparing and contrasting how GVQA's intermediate outputs can help explain successes and failures (and thus, enabling targeted improvements) which is not possible to do for SAN and most other existing VQA models. See \hyperref[sec:appendix_qualitative]{Appendix VIII} for more such examples.

\vspace{-5pt}
\section{Conclusion}
\label{sec:conclusion}
\vspace{-5pt}
\model is a first step towards building models which are visually grounded by design. Future work involves developing models that can utilize the 
best of both worlds (visual grounding and priors), such as, answering a question based on the knowledge about the priors of the world (sky is usually blue, grass is usually green) when the model's confidence in the answer predicted as result of visual grounding is low.

\textbf{Acknowledgements.} We thank Yash Goyal for useful discussions. This work was funded in part by: NSF CAREER awards, ONR YIP awards, Google FRAs, Amazon ARAs, DARPA XAI, and ONR Grants N00014-14-1-\{0679, 2713\} to DB, DP, and PGA Family Foundation award to DP.


\section*{Appendix Overview}
\label{sec:appendix_overview}

\noindent In the appendix, we provide:
\vspace{-10pt}
\begin{enumerate}[I]
\setlength{\itemsep}{1pt}
\setlength{\parskip}{0pt}
\setlength{\parsep}{0pt}
\item - Additional analysis of \cpData splits (\hyperref[sec:appendix_dataset]{Appendix I})

\item - Details of benchmarking VQA models on \cpData (\hyperref[sec:appendix_baselines]{Appendix II})

\item - Implementation details of \model (\hyperref[sec:appendix_model]{Appendix III})

\item - Additional splits of \cpDataNew
(\hyperref[sec:appendix_additional_splits]{Appendix IV})

\item - Performance of model components on \cpDataNew 
(\hyperref[sec:appendix_model_components]{Appendix V})

\item - Performance of SAN with $Q_{main}$
(\hyperref[sec:appendix_san_with_qmain]{Appendix VI})

\item - Performance of GVQA - VCC$_{loss}$ on \origDataOld and \origDataNew (\hyperref[sec:appendix_gvqa_without_vcc_results]{Appendix VII})


\item - Additional qualitative examples (\hyperref[sec:appendix_qualitative]{Appendix VIII})
\end{enumerate}
\section*{Appendix I: Additional analysis of \cpData splits}
\label{sec:appendix_dataset}

\figref{fig:ans_dist} shows the distribution of answers for several question types such as \myquote{what color}, \myquote{what sport}, 
\myquote{how many}, etc. for the train (left) and test (right) splits of the \cpDataNew dataset (the distribution of answers for \cpDataOld is presented in \secref{sec:dataset}). We can see that the distributions of answers for a given question type is significantly different for train and test. For instance, \myquote{tennis} is the most frequent answer for the question type \myquote{what sport} in \cpDataNew train split whereas \myquote{baseball} is the most frequent answer for the same question type in \cpDataNew test split. 
Similar differences can be seen for other question types as well -- \myquote{does}, \myquote{which}.

\begin{figure}[h]
\centering
\includegraphics[width=1\linewidth]{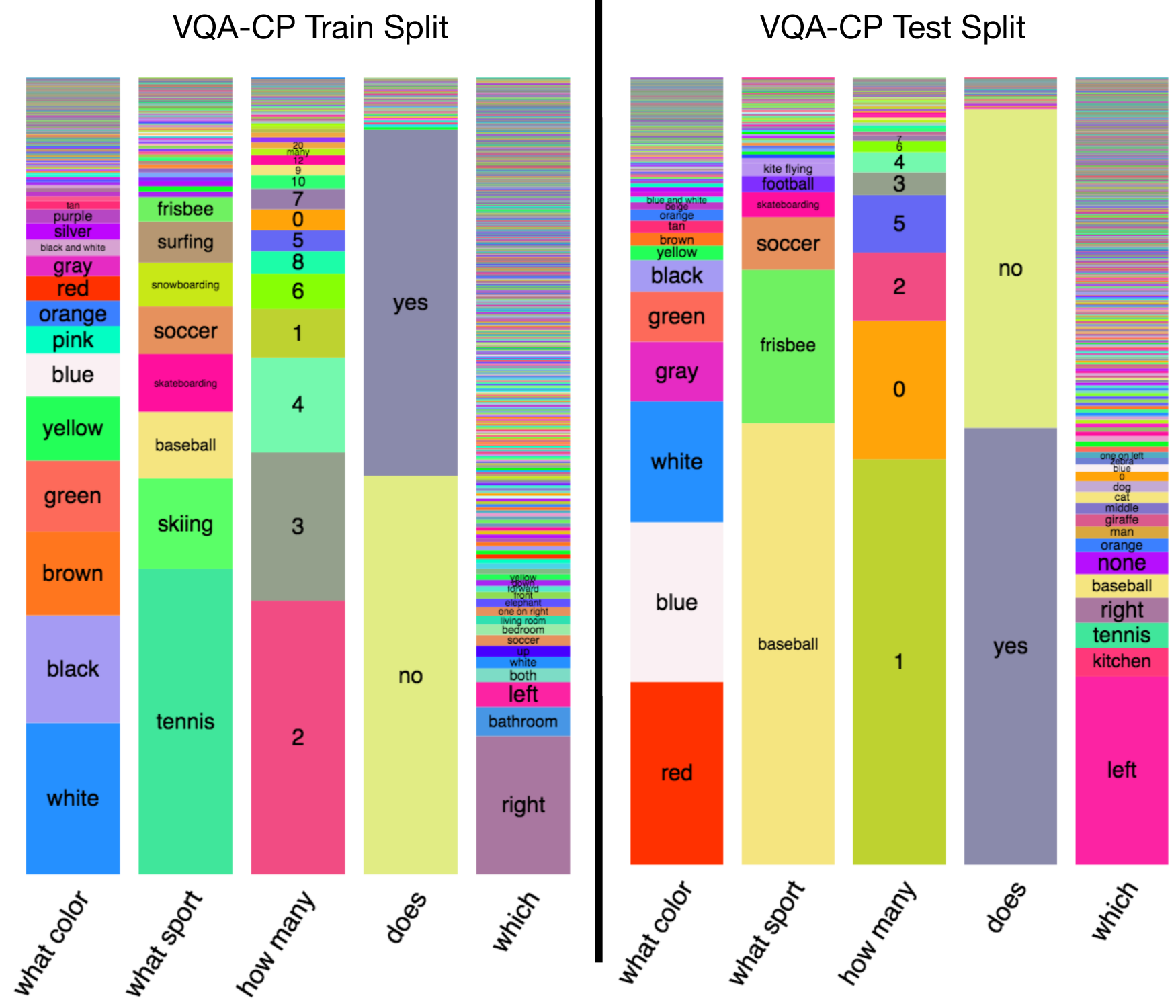}
\caption{Distribution of answers per question type vary significantly between \cpDataNew train (left) and test (right) splits. For instance, \myquote{white} and \myquote{black} are commonly seen answers in train for \myquote{What color}, where as \myquote{red} is the most frequent answer in test. These have been computed for a random sample of 60K questions.}
\label{fig:ans_dist}
\end{figure}
\section*{Appendix II: Details of benchmarking VQA models on \cpData}
\label{sec:appendix_baselines}
Below we provide brief descriptions of all the existing VQA models used for benchmarking on \cpData splits:

\textbf{Deeper LSTM Question (d-LSTM Q)} \cite{VQA}: Predicting the answer using question alone (``blind'' model). It encodes the question using an LSTM and passes the encoding through a Multi-Layered Perceptron (MLP) to classify into answers.

\textbf{Deeper LSTM Question + normalized Image (d-LSTM Q + norm I)} \cite{VQA}: The baseline VQA model. This model consists of a Multi-Layered Perceptron (MLP) fed in by normalized image embeddings (produced by VGG-Net~\cite{VGG}) and question embeddings (produced by a 2 layered LSTM). The MLP produces a distribution over top 1000 answers.

\textbf{Neural Module Networks (NMN)} \cite{DBLP:journals/corr/AndreasRDK15}: The model designed to be compositional in nature. The model consists of composable modules where each module has a specific role (such as detecting a dog in the image, counting the number of dogs in the image, etc.). Given an image and the natural language question about the image, NMN decomposes the question into its linguistic substructures using a parser to determine the structure of the network required to answer the question.

\textbf{Stacked Attention Networks (SAN)} \cite{DBLP:journals/corr/YangHGDS15}: One of the widely used models for VQA. Given an image and question, SAN uses the question to attend over the image, using a multi-hop architecture.\footnote{We use a torch implementation of SAN, available at \url{https://github.com/abhshkdz/neural-vqa-attention}, for our experiments.}

\textbf{Multimodal Compact Bilinear Pooling (MCB)} \cite{fukui}: The winner of the VQA Challenge (on real images) 2016. MCB uses multimodal compact bilinear pooling to predict attention over image features and also to combine the attended image features with the question features.

\textbf{Question-type trends of model performance on \cpData:} Examining the accuracies of the above VQA models for different question types shows that the performance drop from \origData to \cpData is larger for some question types than the others. For \cpDataOld, all the models show a significant drop ($\sim$70\%) for \myquote{is there a} questions (such as \myquote{Is there a flowering tree in the scene?}). For such questions in the \cpDataOld test split, the correct answer is \myquote{yes} whereas the prior answer for questions starting with \myquote{Is there a} in \cpDataOld train split is \myquote{no}. So, models tend to answer the \cpDataOld test questions with \myquote{no} driven by the prior in the training data. Some other examples of question types in \cpDataOld resulting in significant drop in performance (more than 10\%) for all the models are -- \myquote{is this an}, \myquote{do you}, \myquote{are there}, \myquote{how many people are}, \myquote{what color is the}, \myquote{what sport is}, \myquote{what room is}, etc. Examples of question types in \cpDataNew resulting in more than 10\% drop in performance for all the models are -- \myquote{is it}, \myquote{is he}, \myquote{are there}, \myquote{how many people are in}, \myquote{what color is the}, \myquote{what animal is}, \myquote{what is in the}, etc.
\section*{Appendix III: Implementation details of \model}
\label{sec:appendix_model}

For the Question Classifier, we use a single layer LSTM with 512$d$ hidden state and train it using the binary cross-entropy loss. For the Answer Cluster Predictor (ACP), we use a single layer LSTM with 256$d$ hidden state and train it using the cross-entropy loss (cross-entropy over 50 classes, corresponding to 50 answer clusters). For the Visual Concept Classifier (VCC), we use a single layer LSTM with 512$d$ hidden state to encode Q$_{main}$, the VGG-Net \cite{VGG} to extract the activations of the last pooling layer (514 x 14 x 14) and the 2-hop attention architecture from SAN \cite{DBLP:journals/corr/YangHGDS15}. We use the binary cross-entropy loss to train each classifier in the VCC. For a given training instance, only a subset of all concept clusters are activated, and only these activated clusters contribute to the loss.

For the Question classifier, the ACP and the VCC, we use the rmsprop optimizer with a base learning rate of 3e-4. For the Answer Predictor (AP) and the Visual Verifier (VV), we use the Adam optimizer with a base learning rate of 3e-3 and 3e-4 respectively. All the implementation is using the torch deep learning framework \cite{torch}. 


\textbf{Effect of number of clusters, clustering algorithm, POS tagger:} As mentioned in \secref{sec:model}, we used 50 clusters and K-means clustering algorithm for clustering the answer classes for the Answer Cluster Predictor (ACP). We tried 25 and 100 clusters as well and found that changing the number of clusters in K-means from 50 to 25 results in 1.05\% drop, from 50 to 100 results in 0.76\% drop in the overall VQA accuracy for the VQA-CP v2 dataset. We also tried Agglomerative clustering (instead of K-means) and found that it results in 0.42\% drop in the overall VQA accuracy on the VQA-CP v2 dataset. Finally, we tried using Spacy POS tagger (instead of NLTK) for the Concept Extractor (CE) and found that it results in 0.02\% improvement in the overall VQA accuracy on the VQA-CP v2 dataset.
\section*{Appendix IV: Additional splits of \cpDataNew}
\label{sec:appendix_additional_splits}

\vspace{-7pt}
\begin{figure}[h]
\centering
\includegraphics[width=\linewidth]{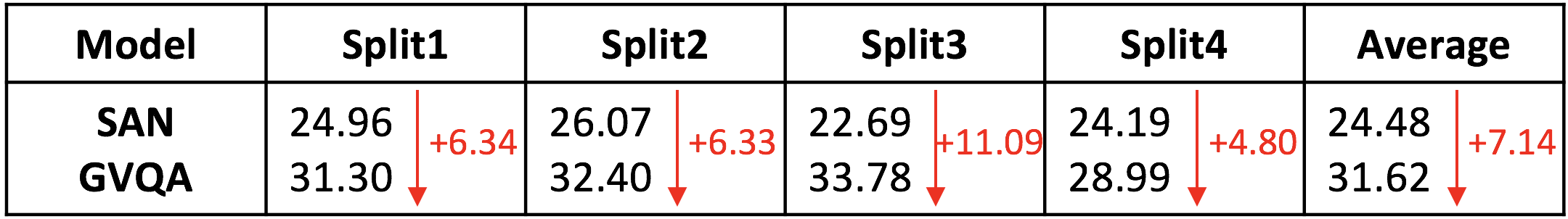}
\caption {Performance of SAN and GVQA for different VQA-CP v2 splits. GVQA consistently outperforms SAN across all splits.}
\vspace{-15pt}
\label{fig:diff_splits}
\end{figure}
\vspace{5pt}

As mentioned in \secref{sec:cp_exp}, to check if our particular VQA-CP split was causing some irregularities in performance, we created three additional sets of VQA-CP v2 splits with different random seeds. We evaluated both SAN and GVQA on all four splits (please see \figref{fig:diff_splits}). We can see that GVQA consistently outperforms SAN across all four splits with average improvement being 7.14\% (standard error: 1.36).

\section*{Appendix V: Performance of model components on \cpDataNew}
\label{sec:appendix_model_components}

\textit{Question Classifier}: On the \cpDataNew test set, the LSTM based question classifier obtains 99.30\% accuracy. 
\textit{ACP}: The Top-1 test accuracy is 51.33\%, with 80.12\% for questions whose answers are in attribute clusters and 39.21\% for questions whose answers are in object clusters. The Top-3 accuracy rises to 63.22\%. Note that these accuracies are computed using the automatically created clusters.
\textit{VCC}: The weighted mean test F1 score across all classifiers is 0.53. The individual concepts are weighted as per the number of positive samples, reflecting the coverage of that concept in the test set.
\section*{Appendix VI: Performance of SAN with $Q_{main}$}
\label{sec:appendix_san_with_qmain}

\vspace{-7pt}
\begin{table}[h]
\setlength{\tabcolsep}{3pt}
{\small
\begin{center}
\begin{tabular}{@{}lcccc@{}}
\toprule
Model & Overall & Yes/No & Number & Other \\
\midrule
SAN \cite{DBLP:journals/corr/YangHGDS15} & 24.96 & 38.35 & 11.14 & 21.74 \\
SAN - $Q_{full}$ + $Q_{main}$ & 26.32 & 44.73 & 09.46 & 21.29 \\
\model\ (Ours) & \textbf{31.30} & \textbf{57.99} & \textbf{13.68} & \textbf{22.14} \\
\bottomrule
\end{tabular}
\end{center}
}
\vspace{-10pt}
\caption {Performance of SAN - $Q_{full}$ + $Q_{main}$ compared to SAN and GVQA (our model) on \cpDataNew dataset. GVQA outperforms both SAN and SAN - $Q_{full}$ + $Q_{main}$.}
\vspace{-15pt}
\label{table:san_with_qmain}
\end{table}
\vspace{5pt}

As mentioned in \secref{sec:gvqa_components}, as an additional check, we trained a version of SAN where the input is $Q_{main}$ instead of $Q_{full}$. \tableref{table:san_with_qmain} shows the results of this version of SAN (SAN - $Q_{full}$ + $Q_{main}$) along with those of SAN and GVQA on VQA-CP v2. We can see that this version of SAN performs 1.36\% (overall) better than the original SAN, however still 4.98\% (overall) worse than GVQA (with $Q_{main}$).
\section*{Appendix VII: Performance of GVQA - VCC$_{loss}$ on \origDataOld and \origDataNew}
\label{sec:appendix_gvqa_without_vcc_results}

\begin{table*}[h!]
\setlength{\tabcolsep}{5pt}
{\small
\begin{center}
\begin{tabular}{@{}l@{\hspace{4\tabcolsep}}lccc@{\hspace{4\tabcolsep}}cccc@{}}
\toprule
& \multicolumn{4}{c}{\origDataOld} & \multicolumn{4}{c}{\origDataNew} \\
\cmidrule{2-9}
Model & Overall & Yes/No & Number & Other & Overall & Yes/No & Number & Other\\
\midrule
SAN & 55.86 & 78.54 & 33.46 & 44.51 & 52.02 & 68.89 & 34.55 & 43.80 \\
GVQA - VCC$_{loss}$ & 48.51 & 65.59 & 32.67 & 39.71 & 48.34 & 66.38 & 31.61 & 39.05 \\
GVQA & 51.12 & 76.90 & 32.79 & 36.43 & 48.24 & 72.03 & 31.17 & 34.65 \\
\midrule
Ensemble (SAN, SAN) & 56.56 & 79.03 & 34.05 & 45.39  & 52.45 & 69.17 & 34.78 & 44.41 \\
Ensemble ((GVQA - VCC$_{loss}$), SAN) & 56.44 & 78.27 & 34.45 & 45.62 & 51.79 & 68.59 & 34.44 & 43.61 \\
Ensemble (GVQA, SAN) & 56.91 & 80.42 & 34.40 & 44.96 & 52.96 & 72.72 & 34.19 & 42.90 \\
\midrule
Oracle (SAN, SAN) & 60.85 & 83.92 & 39.43 & 48.96 & 56.68 & 74.37 & 40.08 & 47.61 \\
Oracle ((GVQA - VCC$_{loss}$), SAN) & 64.47 &  90.17 & 42.92 & 50.64 & 61.93 & 85.13 & 43.51 & 49.16 \\
Oracle (GVQA, SAN) & 63.77 & 88.98 & 43.37 & 50.03 & 61.96 & 85.65 & 43.76 & 48.75 \\
\bottomrule
\end{tabular}
\end{center}
}
\caption {Results of GVQA, GVQA - VCC$_{loss}$ and SAN on \origDataOld and \origDataNew when trained on the corresponding train splits. Please see text for more details.}
\label{table:appendix_vqa_orig}
\end{table*}


\tableref{table:appendix_vqa_orig} presents the full results (i.e., broken down into Yes/No, Number and Other) of three models -- \model, \model - VCC$_{loss}$ and SAN, along with their ensembles and Oracle performances. 
We can see that \model - VCC$_{loss}$ performs worse than \model on \origDataOld and similar to \model on \origDataNew. So in addition to interpretability, \model is overall better than \model - VCC$_{loss}$ on these original VQA splits. Another observation about \model - VCC$_{loss}$ is that the Oracle ((\model - VCC$_{loss}$), SAN)'s overall performance is 8.61\% higher than that of SAN for \origDataOld (9.91\% for \origDataNew), suggesting that \model - VCC$_{loss}$ has strengths complementary to SAN (just like \model). Note that Oracle ((\model - VCC$_{loss}$), SAN) is higher than Oracle (SAN, SAN) for both \origDataOld and \origDataNew, suggesting that \model - VCC$_{loss}$'s complementary strengths are more than that of another SAN model (with a different random initialization). Inspired by this, we report the performance of the ensemble of \model - VCC$_{loss}$ and SAN ((\model - VCC$_{loss}$) + SAN) in \tableref{table:appendix_vqa_orig}, where the ensemble combines the outputs from the two models using product of confidences of each model. Unlike \model + SAN, (\model - VCC$_{loss}$) + SAN does not outperform SAN + SAN (worse by 0.12\% overall for \origDataOld and by 0.66\% overall for \origDataNew). Hence, \model is a better complement of SAN than \model - VCC$_{loss}$, in addition to being more transparent.
\section*{Appendix VIII: Additional qualitative examples}
\label{sec:appendix_qualitative}

\begin{figure}[h]
\centering
\includegraphics[width=1\linewidth]{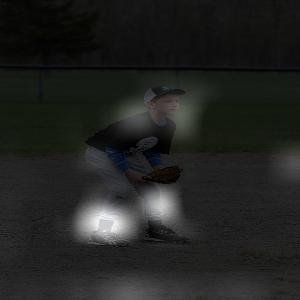}
\caption{VCC's attention map for the example shown in \figref{fig:qual2} (right)}
\label{fig:att_map}
\end{figure}

\figref{fig:att_map} shows the VCC's attention map for the example shown in \figref{fig:qual2} (right). Please refer to \figref{fig:qual2} for more details.

\begin{figure*}[h]
\centering
\includegraphics[width=0.9\linewidth]{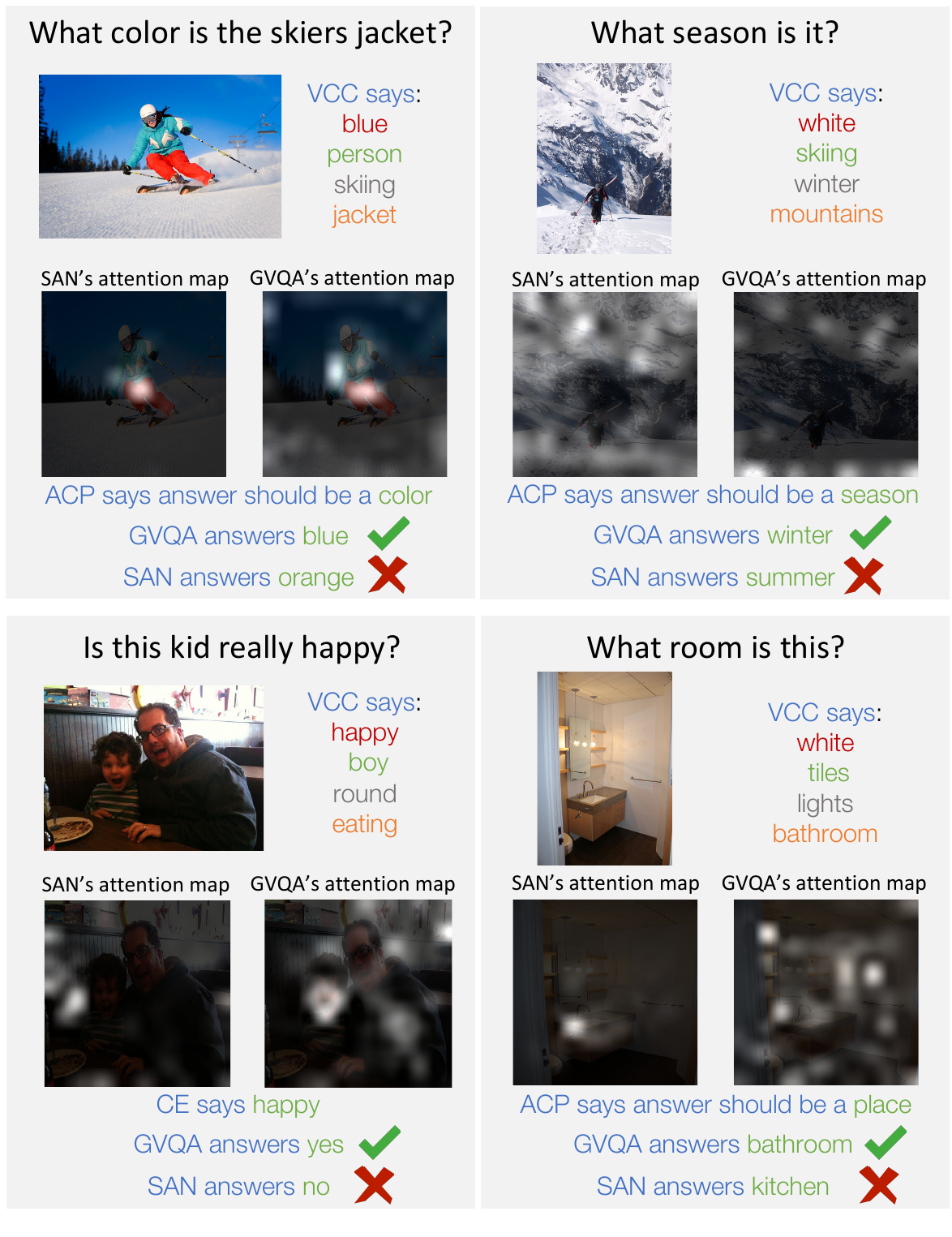}
\caption{\textbf{Transparency of GVQA.} For each of the above examples, GVQA's intermediate predictions can help explain why it predicted what it predicted. \textbf{Top-left:} VCC predicts the following visual concepts -- blue, person, skiing and jacket. ACP predicts the cluster corresponding to colors. Finally, GVQA predicts \myquote{blue} as the answer. So, we can see why GVQA predicts \myquote{blue} -- because, of all the visual concepts predicted by VCC, only \myquote{blue} represents a color. Looking at the attention maps can further indicate why GVQA is ``seeing'' blue (because it is ``looking'' at the jacket as well, unlike SAN which is only ``looking'' at the pants). SAN's prediction is \myquote{orange} and unlike GVQA, SAN's architecture does not facilitate producing such an explanation, which makes it difficult to understand why it is saying what it is saying. \textbf{Top-right}: Both GVQA and SAN are ``looking'' at the regions covered with snow, but GVQA correctly predicts \myquote{winter}, whereas SAN incorrectly predicts \myquote{summer} which is unclear why. \textbf{Bottom-left}: The Concept Extractor (CE) predicts \myquote{happy} whose visual presence is verified by VCC which is ``looking'' at the region corresponding to the kid's face. \textbf{Bottom-right}: GVQA focuses on a larger part of the scene and correctly recognizes it as \myquote{bathroom}.}
\label{fig:qual_pos}
\end{figure*}

\begin{figure*}[h]
\centering
\includegraphics[width=0.9\linewidth]{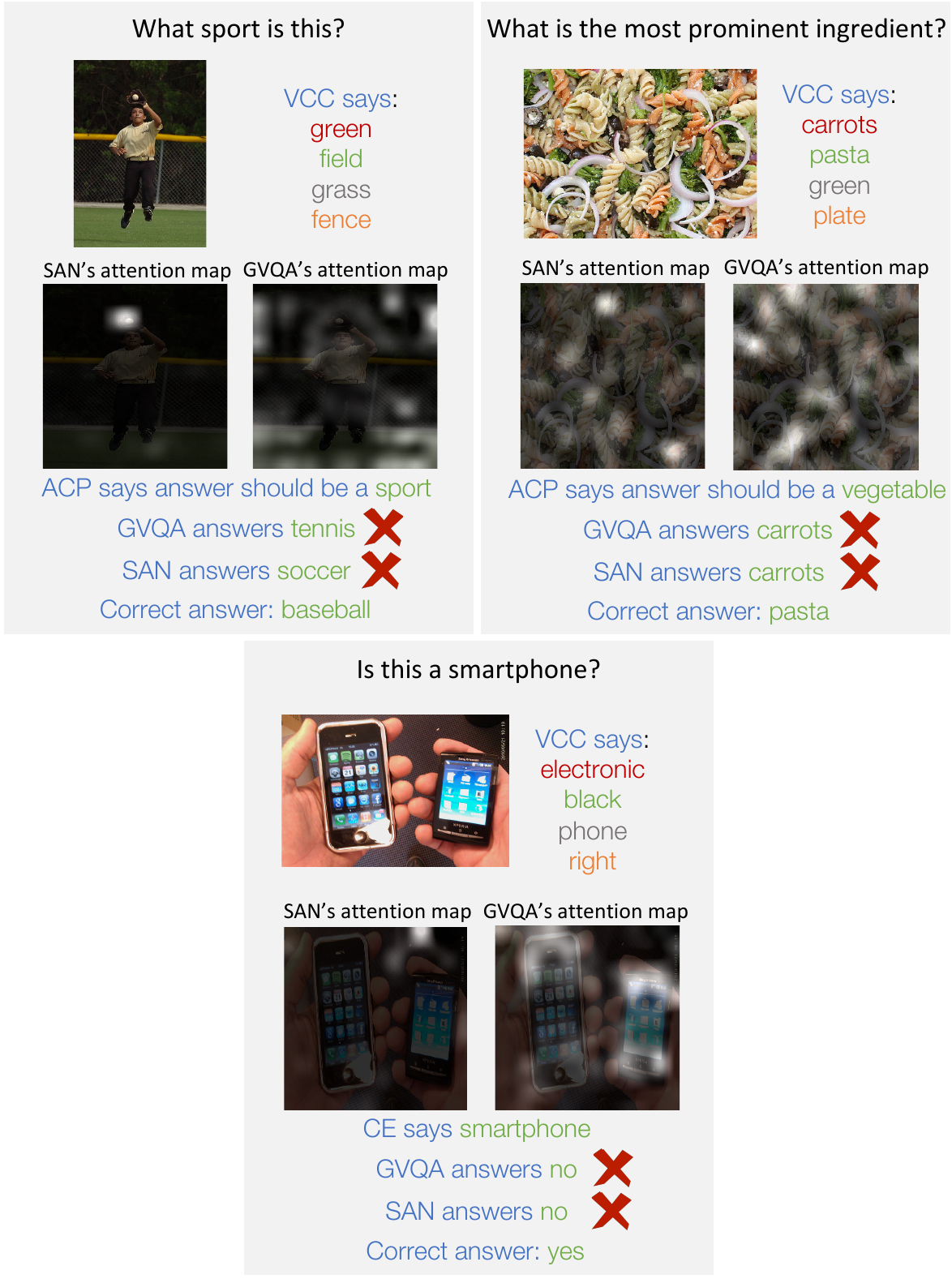}
\caption{\textbf{Transparency of GVQA.} For the above examples, both GVQA and SAN incorrectly answer the question. However, GVQA's intermediate predictions can help explain why it is incorrect. \textbf{Top-left:} For GVQA, VCC's predictions indicate that it is perhaps ``looking'' at the field, which can be further verified by the attention map. SAN's attention map suggests that it is ``looking'' at the ball but still does not explain why it is predicting \myquote{soccer}. Perhaps, it is confusing the ball with a soccer ball. \textbf{Top-right:} The attention maps from GVQA and SAN look similar to each other. However, looking at ACP's and VCC's prediction (for GVQA) suggest that it is indeed ``seeing'' \myquote{pasta} (the correct answer), but still predicting \myquote{carrots} because the ACP is incorrectly predicting the cluster corresponding to vegetables instead of the cluster corresponding to pasta. \textbf{Bottom:} GVQA is ``looking'' at the smartphone (unlike SAN), but yet incorrectly answers \myquote{no}, because the VCC does not recognize the phone as a smartphone. It however correctly predicts \myquote{phone}, \myquote{electronic}, \myquote{black} and \myquote{right}.}
\label{fig:qual_neg}
\end{figure*}

\figref{fig:qual_pos} and \figref{fig:qual_neg} show some qualitative examples from the \cpDataNew test set along with GVQA's and SAN's predicted answers. Also shown are the intermediate outputs from GVQA which provide insights into why GVQA is predicting what it is predicting and hence enable a system designer to identify the causes of error. This is not easy to do in existing VQA models such as SAN.

\clearpage
{\small
\bibliographystyle{ieee}
\bibliography{strings,references}
}
\end{document}